# Relational Learning Analysis of Social Politics using Knowledge Graph Embedding

Bilal Abu-Salih[1,2], Marwan Al-Tawil[1], Ibrahim Aljarah[1], Hossam Faris[1], Pornpit Wongthongtham[2]

[1] *KASIT, The University of Jordan, Amman, Jordan*

[2] *Curtin University, Perth, Australia*

## Abstract

Knowledge Graphs (KGs) have gained considerable attention recently from both academia and industry. In fact, incorporating graph technology and the copious of various graph datasets have led the research community to build sophisticated graph analytics tools. Therefore, the application of KGs has extended to tackle a plethora of real-life problems in dissimilar domains. Despite the abundance of the currently proliferated generic KGs, there is a vital need to construct domain-specific KGs. Further, quality and credibility should be assimilated in the process of constructing and augmenting KGs, particularly those propagated from mixed-quality resources such as social media data. This paper presents a novel credibility domain-based KG Embedding framework. This framework involves capturing a fusion of data obtained from heterogeneous resources into a formal KG representation depicted by a domain ontology. The proposed approach makes use of various knowledge-based repositories to enrich the semantics of the textual contents, thereby facilitating the interoperability of information. The proposed framework also embodies a credibility module to ensure data quality and trustworthiness. The constructed KG is then embedded in a low-dimension semantically-continuous space using several embedding techniques. The utility of the constructed KG and its embeddings is demonstrated and substantiated on link prediction, clustering, and visualisation tasks.

## 1. Introduction

Triggered by the qualitative leap taken by Google in 2012 to coin the so-called Knowledge Graph (KG)[1] as well as the tremendous spread of using Semantic Web technologies, KGs have now been widely incorporated in industry and academia as being a factual reflection to the human knowledge to solve several domain-dependent real-life problems [2]. In fact, the proliferation of Social Big Data has prompted the necessity for sophisticated approaches to assist the machine to better understand the context of the multimodal contents. In particular, the heterogeneity in data sources and format, the discrepancy in vocabulary, and the lack of a comprehensive and fused knowledge island are the key challenges for analysts. Yet, by presenting the domain of knowledge as a set of entities and relations, KGs facilitate constructing a unified standard representation for fusion of data. This thereby has led to knowledge propagation embodying graph datasets of divergent and interrelated domains [3] and has extended to benefit large scale applications such as question answering [4], recommenders systems [5] KG completion [6], entity disambiguation [7], and text classification[8]. As a result, analysts today are able to conduct an in-depth analysis of external business data such as customer blog postings [9], Internet chain-letter data [10], social tagging [11], Facebook news feed [12] and many other semantic Artificial Intelligence applications.

Despite the widespread usage of domain-independent (open-world) KGs such as Google KG, Freebase[1], YAGO[2], Dublin Core (DC[3]), Simple Knowledge Organization System (SKOS[4]), Semantically-Interlinked Online Communities (SIOC[5]), and DBPedia[6] knowledge base, the domain-dependent KGs provide an overabundance of benefits to tackle domain-specific problems as well as to gain the hoped-for added value from domain corpora [13]. Domain knowledge is commonly captured in a KG, which is then used to enrich the semantics of data with a specific conceptual representation of entities. The reuse of domain ontology and interlinking process of embodied classes, entities, and concepts with other relevant entities from other KG repositories, facilitates the interoperability of information. Through leveraging domain ontology and semantic web tools, KGs enable constructing conceptual hierarchies and populating the domain ontology with instances extracted using knowledge extraction techniques. Hence, KGs are used as backbones to support intelligent systems by extracting the semantics of textual data that is collected from different vocabularies and semantic repositories to enrich the semantic description of resources using an annotation component.

The ongoing efforts to construct large-scale KGs have been notably increasing. This has led to produce massive KGs embodying billions of facts that describe different contexts [14]. These KGs, however, suffer from incompleteness, which negatively affects the utility of such graphs to be leveraged in real-life applications [15]. For example, Freebase, a large-scale KG and commonly used knowledge-base in research communities, is far from completeness; it has been indicated that the "place of birth" for above 70 percent of "Person" entities are missing, and more than 90 percent of the person entities on freebase have no embodied ethnicity [16, 17]. This also applies to Wikipedia and many other knowledge bases. This has led the research community to confront this issue by providing technical solutions to tackle it, commonly known as KG Augmentation/Completion approaches. KG Completion (a.k.a Link Prediction) aims to amplify the KG with new facts that are depicted by new likelihood entities and/or new relations. Link prediction has many applications, such as predicting new friendships in social networks and recommender systems to various other use cases. In this context, a new cohort of models has recently gained considerable attention. These models are designed to embed the constituents of a KG (entities and relationships) into a low-dimension semantically-continuous space [18]. The generated embeddings can be then leveraged to generate a set of candidate facts to fulfil a completion task [19].

Another important consideration is the factuality and credibility of the embodied knowledge in a KG. The rapid growth in KGs sizes has risen a question on the quality of the embodied knowledge (i.e. entities and relations), and whether these facts do factually represent the intended real-world entities and interlinked via their relationships. This has posed several challenges, which have driven research in this field. For example, due to the huge amount of information flowing from Online Social Networks (OSNs) to its recipients, in conjunction with the lack of a gatekeeper for those sites, it is difficult to verify the content and thus makes it easier for others to perform the task of disinformation [20]. On the other hand, the good quality content obtained from social media has several significant impacts [21]. Hence, the data collected from OSNs should be examined to augment KGs with trustworthy facts to benefit real-life applications. Although the significant efforts attempted to address quality, endeavours in this direction are inadequate, and several measures should be proposed and taken to maintain the quality of KGs.

This paper presents a novel credibility-based domain-specific KG Embedding (KGE) framework. This framework involves capturing real-life entities obtained from social data into a formal and fused representation depicted by a domain ontology. The proposed approach makes use of various cross-domain knowledge-based repositories including Google KG™, IBM Watson NLU™, and Wordnet™ to enrich the semantics of the textual contents, thereby facilitating the interoperability of information. The proposed framework also embodies a credibility module to ensure data quality and trustworthiness. The constructed KG is then embedded in low dimensional vector space using several embedding techniques. The resultant KG embeddings are used to conduct several tasks including link prediction, clustering, and visualisation. Evaluation protocol and metrics are used to compute the performance of the incorporated embedding models.

---

[1] https://developers.google.com/freebase
[2] http://www.foaf-project.org/
[3] https://www.dublincore.org/
[4] http://www.w3.org/2004/02/skos/
[5] http://sioc-project.org/
[6] https://wiki.dbpedia.org/

In this paper, we have made the following key contributions:

- A domain knowledge graph is constructed based on an extended politics domain augmented ontology using dissimilar light-weight ontologies and semantic repositories.
- An embedded social credibility module is incorporated and customised to enhance the quality of the collected datasets.
- Various state-of-the-art embedding models are implemented and their performance is evaluated using key evaluation metrics.
- The utility of the constructed KG Embeddings is demonstrated and substantiated on link prediction, clustering, and visualisation tasks.

This paper is organised as follows: Section 2 provides background on works related to the context of this paper. Section 3 discusses the overall methodology on proposed framework of this paper and the included modules. The experiments carried out in this study are explained in Section 4 along with the evaluation mechanism and the implemented tasks. Finally, the conclusions and some possible research directions are reported in Section 5.

## 2. Related Works

**Knowledge Acquisition (Completion, Entity and Relation Extraction):** KGs are commonly constructed from (semi-)structured, such as Wikipedia or unstructured datasets, such as web data using Natural Language Processing (NLP) and linguistic approaches as well as other statistical techniques [22]. KG can be further extended and thereby its embedded knowledge can be augmented to include missing facts of the real world. The process of knowledge acquisition can be categorised into two key dimensions, namely: KG completion, and entity and relation extraction. KG completion aims to expand the current knowledge by accumulating more facts to the current state of the KG, while the latter dimension aims to infer new knowledge by predicting new relations and entities [2]. Link and entity inference in the context of KGs is the process of amplifying the KG with new facts depicted by new entities and/or new relations.

Several approaches have been introduced to tackle this issue [6, 23-25] [26]. These attempts have also extended to address interrelated domains [27]. Authors of [23] proposed joint representation learning framework attempts to solve the complexity of the structure of semantic information by presenting a mutual attention mechanism, which can be used to highlight the important features by conjoining the textual content and the KG models. Augmenting knowledge in disaster situations has been also addressed in the literature. For example, Purohit et al. [24] proposed DisasterKG; a disaster KG offers a platform that provides resources to answer critical inquiries. Authors made their point on how interoperability of information from dissimilar data resources can efficiently improve decision making in such cases. Completion of KG that using web pages was attempted by Kruit et al. [28]. Authors of [28] suggested a new approach for HTML table interpretations, where the row and column indicate an entity and an attribute respectively. By using Probabilistic Graphical Model (PGM), authors were able to infer new facts for KGs with dissimilar topologies. Shi and Weninger [29] proposed ConMask; an Open-World Knowledge Graph Completion system. This system is designed incorporating fully convolutional neural networks, and semantic averaging to be able to tackle the incompleteness of KG. The proposed system has proven ability to forecast relations including unseen entities.

**NLP Applications using KGE:** Incorporating graph technology and the abundance of dissimilar graph datasets have assisted in building quite sophisticated graph analytics tools. Despite the effectiveness of the conventional graph analysis approaches, such as Graphx [30], Gephi [31], GraphLab [32] to name a few, graph embedding has notably improved the efficiency of conducting graph analytics aby converting the graph to a low semantic dimensional space, thus information can be represented as vectors leading to computational efficiency. Several efforts have been conducted to incorporate KG Embeddings to address numerous NLP challenges. For example, Yao et al [33] proposed a topic distillation approach embodying Latent Dirichlet Allocation (LDA) to improve document presentation in the semantic space. Authors of [34] benefited from the architecture of a neural network and a constructed knowledge base to build Text Concept Vector (TCV) that can be used to infer high-level presentation of concepts from the textual content. KGs are also utilised in conjunction with deep learning models to distil knowledge for several applications, such as sentiment tasks [35], bilingual dictionary induction [36], fake news detection [37], recommender systems [38], and other miscellaneous applications [39, 40].

**Classification and clustering using KGE:** Classification in the context of KGs is the task of determining, whether the entities/nodes, relations/edges, or the whole triple contained within the testing dataset are correct. This task can be perceived as a binary classification task involving class labelling to each entity, relation or triple. Underneath this broad classification umbrella, quite a few conducted literature reported attempts to efficient and reliable applications incorporating graph embedding [41, 42] and also using dissimilar embedding techniques such as TransR [6], HolE [43] and ANALOGY [44]. On the other hand, clustering is an unsupervised learning approach that aims to assemble similar entities together in groups. Clustering can also be used to examine the efficiency of the approach used for KG embedding. Incorporating KGE boosts the traditional clustering algorithms by transforming the embedded components of the graph into vectors [45]. Other unconventional approaches have been also presented in the literature. For example, Tian et al. [46] showed how utilising deep neural networks can improve KG clustering through mapping the similarity matrix of the input graph to the output graph embedding using the layer-wise pre-training scheme.

## 3. Methodology

### 3.1 Overall Framework Architecture

Figure 1 shows the proposed KGE framework. As depicted in the figure, the system comprises five core components, namely: Domain Knowledge Acquisition & Pre-processing; Domain Knowledge Inference; Knowledge Credibility Module; Knowledge Graph Creation and Embedding; and Knowledge Reasoning. The system collects its dataset from three main knowledge resources, namely: Twitter, Wikipedia, and miscellaneous news articles.

The collected datasets are pre-processed in order to ensure data cleansing and integration, then domain knowledge being captured in domain ontologies is identified and used in the enrichment of the semantics of the textual contents. This process is attained through the domain knowledge inference module - semantic kitchen. This module incorporates several knowledge-based repositories including Google KG™, IBM Watson NLU™, Politics domain Ontology, and Wordnet™. The next phase in this framework is to ensure the credibility of the incorporated knowledge. Credibility of knowledge is commonly neglected in the construction of KGs especially when the knowledge is attained from social media where spammers and other low trustworthy users find a fertile medium to publish and spread their content taking advantage of the open environment and fewer restrictions of these platforms. The following module constructs the domain KG and conducts the KG embedding. This facilitates the knowledge reasoning which is carried out in the last module and represented by incorporated neural machine learning models for relational learning. Details on the system framework and the embodied modules are discussed in the next sections.

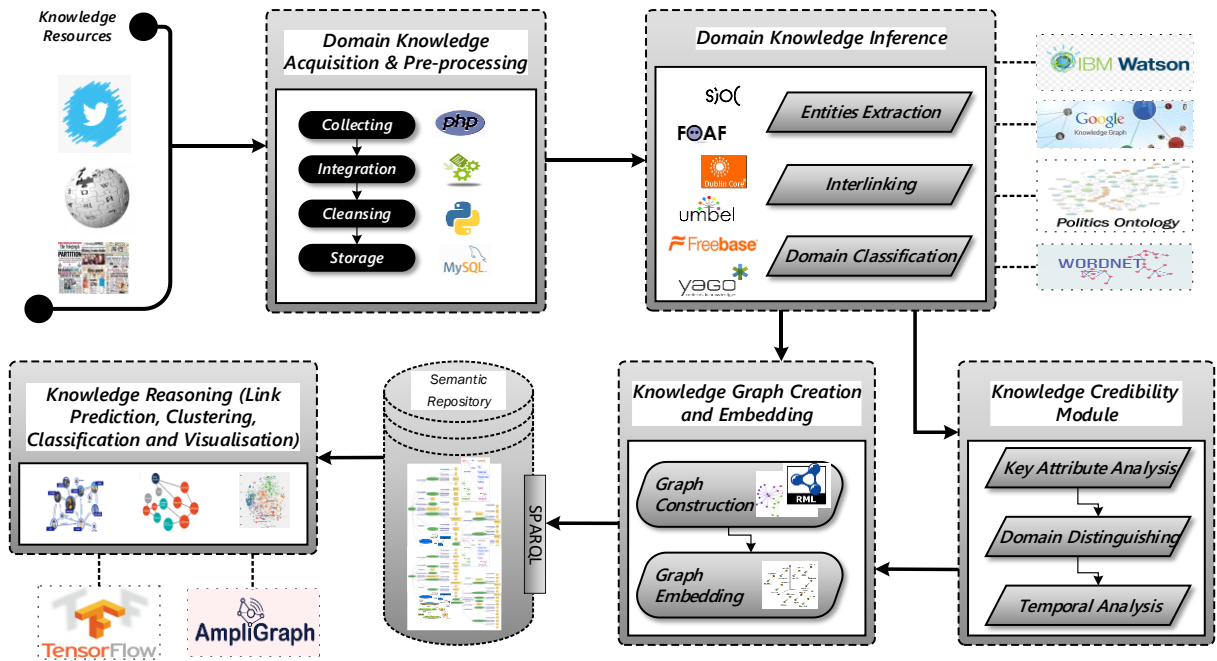

*Figure 1: System Architecture*

## 3.2 Domain Knowledge Acquisition and Pre-processing

Since the emergence of the OSNs, the propagation of social data has revolutionised the research avenues to develop state-of-the-art techniques for social data analytics. OSNs are a fertile medium for researchers in diverse disciplines, leveraging the vast volume of content. For the purpose of proof of concept, this study focuses on analysing the political content that can be collected and distilled from the Twitter platform. The politics domain is selected amongst other domains due to the following reasons: (i) Twitter has been intensively incorporated as an important venue by politicians to express and defend their policies, to practice electoral propaganda, and to communicate with their supporters [47]. (ii) Twitter has raised a lot of controversy about its usage as a platform to attack political opponents [48]. (iii) Twitter is characterized by its growing social base to include broad political social groups leveraged by the ease of use, free to access, and less governmental control [49]. (iv) In fact, the amount of the political discourses amongst the overall social content is overwhelming; over one-third of OSN's users believe that they are worn-out by the quantity of the political content they encounter [50]. The social dataset used for this study has been collected using the twitter's "User_timeline[7]" API method. This mechanism allows access to and retrieval of the public users' content and metadata.

### 3.2.1 Dataset Acquisition

This study aims to augment the constructed domain-based knowledge graph with facts attained from heterogeneous data sources. These facts will not be only obtained from politics-related sources, but also be gathered from users who do not explicitly indicate an interest in this designated domain. Further, users who will be potentially detected as spammers will be also included to prove the applicability of our approach to filter out those users, thereby enhancing the quality of imported facts as it will be discussed later.

As for users who explicitly indicate an interest in politics domain are collected from various resources as follows: (i) we gathered all information provided for members listed in Parliament of Australia official website[8]. Those include Senators and members of the Australian House of Representatives. (ii) A selected set of users is assembled from three distinguished Australian Twitter lists that are relevant to the political domain[9]. (iii) Mixt sources[10]. Users whose political interest is not explicitly identified were tentatively selected various Australian Twitter's lists established to discuss sports, Information Technology, and other non-politics domains. Finally, we included a subset of users indicated in the Twitter graph dataset collected by Akcora et al. [51]. This graph was used in experiments carried out by Akcora et al. to discover spammers and other illegitimate accounts. One of the contributions of this paper is to provide a platform where trustworthy social content can be imported to augment the domain KG, and by this means eliminate untrustworthy content. Hence, the reason behind selecting the graph of Akcora et al. [51] is twofold: (i) to proof the efficiency and applicability of the proposed approach which can be used to eliminate spammers and their content and entrench the domain KG with trustworthy facts; (ii) to embed also the content of domain influencers from a dataset of users whose domains of knowledge are not explicitly known.

---

[7] https://dev.twitter.com/rest/reference/get/statuses/user_timeline
[8] http://www.aph.gov.au/

### 3.2.2 Dataset Pre-processing:

One of the significant features toward properly addressing and curating Big data is to ensure its veracity. The veracity of data refers to the certainty, faultlessness, and trustworthiness of data [52]. Although reliability, availably, and security of data's source is significant [52], these factors do not guarantee data correctness and consistency especially in the context of social media where data can be infected with spam and other junk contents. Hence, appropriate data cleansing, integration, and credibility techniques should be incorporated to ensure certainty and veracity of data. The collected users and their contents are cleansed and integrated to enhance quality as follows :

**Datasets cleansing**: cleansing data is a crucial step to improve the quality of data that will be used in further analysis. Hence, detecting and removing errors and corrupted data, meaningless data, redundant data, and irrelevant data are key techniques in data cleansing which are carefully carried out in this experiment to guarantee that only curated data are passed for the next phase.

**Data quality enhancement:** the list of Twitter handles (a.k.a. screen name such as @*username*), which are indicated in the user's metadata, is collected and replaced with the actual user's corresponding name. To achieve this task, Twitter provides a RESTful API service called "lookup[11]" that is used to reveal the twitterer with a certain handle by receiving full hydrated information about the user. Twitter handles are commonly neglected in twitter mining applications. However, handles are used to mention for example twitterers of important entities that are related to a certain domain. For example, a user demonstrates an interest in the political domain if the user is commonly posting politics-related content as well as mentioning twitterers related to politics domain such as politicians or political parties. Hence, it is essential to identify and determine the actual user information of those handles. This assists in the process of domain modelling and inference.

### 3.3 Domain Knowledge Modelling and Inference

Domain knowledge modelling inference is the key phase in the proposed timeline. Knowledge modelling presents the core activity in knowledge graph creation. It involves capturing the real-life entities obtained from the social data into a formal representation depicted by domain ontology. Tom Gruber generated expansive interest across the computer science community by defining ontology as "an explicit specification of a conceptualisation" (Gruber 1993). While conceptualisation aims to formulate the knowledge about real-world entities, the specification attempts to represent those captured entities in a concrete form [53]. Therefore, ontology captures the domain knowledge through the defined concrete concepts (representing a set of entities), constraints, and the relationship between concepts, thereby providing a common understeering of the domain as well as giving a formal representation in machine-understandable semantics. The purpose of an ontology is to represent, share, and reuse existing domain knowledge. This module aims to detect and infer the user's domain of knowledge from pre-processed datasets. For a proof of concept purpose, we experiment within the Politics domain. We use Politics ontology, WordNet, and ontology interoperability and integration to infer the political knowledge.

---

[9] https://twitter.com/latikambourke/lists/australian-journalists/subscribers; https://twitter.com/lizziepops/lists/politics/members; https://twitter.com/smh/lists/federal-politicians
[10] http://earleyedition.com/2009/04/22/australias-top-100-journalists-and-news-media-people-on-twitter; Wikipedia: Australian political journalists : https://en.wikipedia.org/wiki/Category:Australian_political_journalists
[11] https://developer.twitter.com/en/docs/accounts-and-users/follow-search-get-users/api-reference/get-users-lookup

**Politics Ontology:** The BBC offers an array of domain ontologies which are designed to conceptualise a predefined set of domains such as, sports, music, education, to cite a few [54]. These domain ontologies are designed to consolidate the established BBC Linked Open Data platform. Politics domain is amongst the ontologies constructed by BBC and is described as the conceptual knowledge captured in politics ontology along with its embodying knowledge base. BBC defines politics ontology as "an ontology which describes a model for politics, specifically in terms of local government and elections" [55]. Figure 2 displays the BBC Politics ontology. This ontology is initially designed to capture politics in the context of UK government elections.

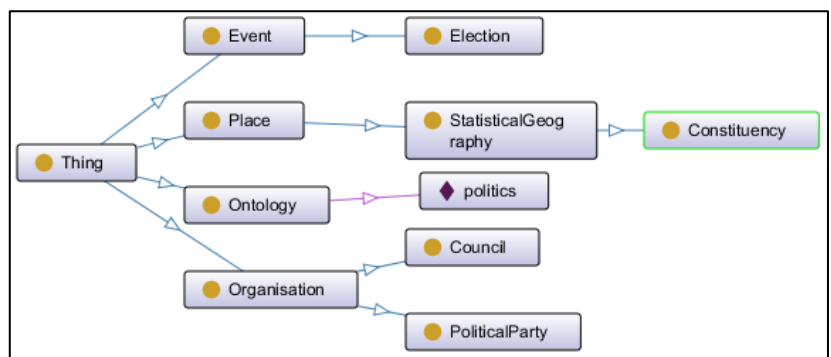

*Figure 2: BBC Politics ontology*

However, the concepts and relationships embedded in the designated ontology are inadequate to properly model this nominated domain, particularly that this study addresses the domain of the politics in the Australian context. Hence, we extend the BBC politics ontology to provide a better depiction to the political domain. Figure 3 shows the extended version of the BBC Politics ontology which is used in this research.

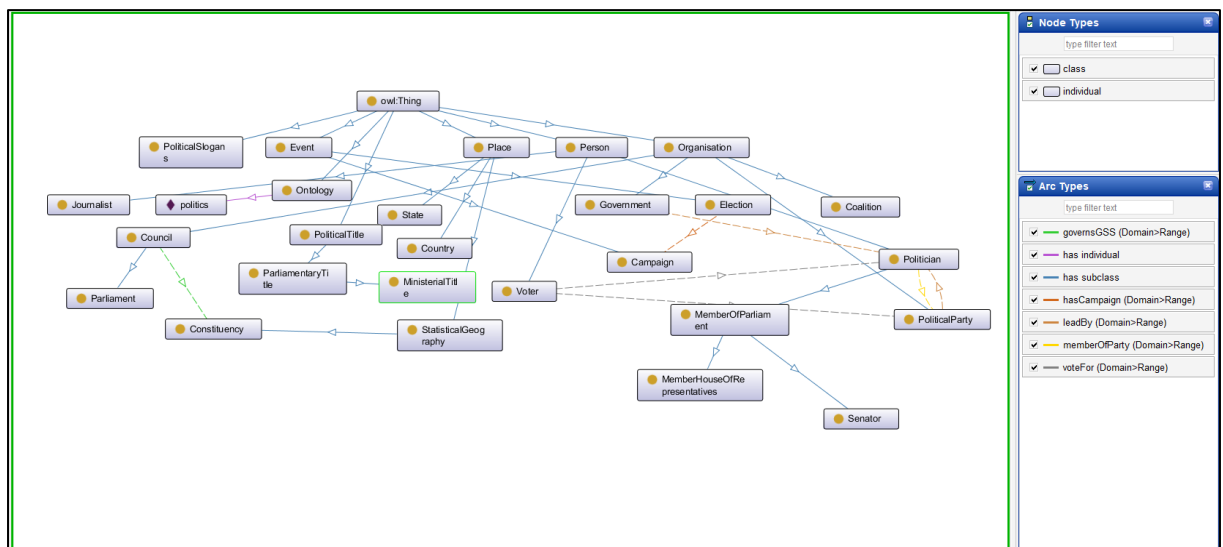

*Figure 3: BBC politics ontology extension*

Designing high-quality ontology is important as a corner store to provide meaningful, contextualised, valid, error-free knowledge base. Also, the developed ontology for any domain should be appropriate to answer queries over its semantic concepts, relationships and instances. The new extended version of

Politics ontology is therefore verified to ensure logical consistency. This has been carried out using reasoning process. In particular, the extended ontology is reasoned incorporating various well-known reasoners such as TrOWL, RacerPro, Pellet, Pellet (Incremental), HermiT, FaCT++. Besides the standard inference services provided by ontology reasoners such as classification and realisation, reasoners are generally used to scrutinise all concepts, properties, and instances and embedded hierarchies. Also, they check if concepts are satisfiable and their descriptions are free of contradictory. The new extended Politics Ontology is reasoned and verified, and no contradictory facts are indicated.

**WordNet Database:** WordNet[12] is a vocabulary lexicon that includes a collection of terms/words(synsets /synonyms) that are interrelated and have similar semantic meanings. WordNet is commonly used to augment the term with further other semantic-related concepts that can enrich its meaning. For example, various synsets of the same contextual meaning can be extracted for the term 'Pol, such as "*politician, politico, and political leader*". WordNet is used in this study to expand the knowledge base with synonyms to concepts inserted in the extended Politics Ontology.

**Ontology Interoperability:** Ontology interoperability aims to align and consolidate the developed ontology with relevant entities captured from other predefined domain and generic ontologies. Ontology interoperability is attained in this study by apprehending equivalent links (URIs) that indicate the same entity/resource. This linkage is depicted by using, for example, owl#sameAs relation for the resources in the Linked Data. This entails that URIs of both subject and object indicate the same resource. For the interlinking process, we incorporate Google KG, a knowledge base that is mainly developed to enhance Google's search engines by providing relevant, semantically-enhanced, and context-specific results. The Google KG Search API[13] is used to infer entities and categorised classes/types. The inferred entities are filtered and those are interlinked with the extended politics ontology are captured and used to enrich the ontology. We also utilised Natural Language Understanding service of IBM Watson™ as a one-stop-shop, leveraging access to a wide variety of linked data resources through providing easy access APIs. These resources include but are not limited to: different vocabularies such as Upper Mapping and Binding Exchange Layer (UMBEL), Freebase which are community-curated databases for well-known people, places, and things, YAGO high-quality knowledge base, etc.

IBM Watson is also used for domain-based classification. In particular, IBM Watson analyses the given text or URL and categorises the content of the text or webpage according to a set of domains (taxonomies) with the corresponding scores. Scores are calculated using IBM Watson, range from "0" to "1", and convey the precise degree of an assigned Category/Taxonomy/Domain to the processed text or webpage. IBM Watson presents an inclusive list of categories divided into certain predefined hierarchies where the high-level category indicates the high-level category and the deeper-level category provides a fine-grain category analysis. For instance, "law, govt and politics" is considered a high-level category in which "presidential elections" is one of its deep-level categories. IBM Watson is used further to identify the overall positive or negative sentiment of the provided content. The taxonomy inference module is used in this research in the domain discovery process, while sentiment analysis is used to discover the sentiments of tweets' replies. The purpose of domain classification and sentiment analysis is discussed in the following section.

### 3.4 Knowledge Credibility Module
As mentioned previously, this study aims to make use of domain-specific politics ontology and available KGs to analyse the social contents of users in OSNs, thereby augmenting the domain KG with facts inferred from users with legitimate and credible interest in politics domain. However, the OSNs medium allows legitimate and genuine users as well as spammers and other low trustworthy users to publish and spread their content leveraging the open environment and fewer restrictions [56-63]. Hence, it is vital to measure users' credibility in numerous domains, therefore define domain-based influential users, and filter out spammers and low trustworthy users.

---
[12] https://wordnet.princeton.edu/
[13] https://developers.google.com/knowledge-graph

This paper incorporates CredSaT [60]; a comprehensive credibility mechanism intended to measure users' credibility based on their domains of knowledge. CredSaT provides an effective solution to discover spammers and influential domain-based users from the list of users whose domain(s) of knowledge is tacit, incorporating the temporal factor. The outcome of the credibility module is a ranked list of users with a corresponding credibility value for each specific domain. The temporal factor is assimilated in CredSat; the dataset of a user's data and metadata is divided into several chunks, where each chunk represents a specific period. A metric of credibility measurements is used to evaluate the user's trustworthiness in each particular chuck, thus providing overall credibility values. The mechanism used to calculate a user's value in each step considers other users' values, thereby providing a normalisation approach for building the relative ranking list of credibility in each domain. Hence, each particular key-value obtained from the user's data and metadata is measured against other users' values. In other words, each of the key attributes is normalised in each domain by dividing the value of the user's attribute by the maximum value achieved by all users in that domain. CredSaT shows the effectiveness of its embodied framework by benchmarking it against other state-of-the-art baseline models.

As mentioned previously our study uses the Twitter graph dataset crawled by Akcora et al. [51]. This dataset comprises spammers and other anomalous users. Hence, the main purpose of the knowledge credibility module is to filter out spammers and other low trustworthy users as their social contents affect the quality of the incorporated domain-based knowledge. For example, spammers who hijack tweets of politics-related contents, event, and stories should be eliminated from further conducted analysis despite the fact that political entities extracted from the contents of those users are relatively high. Table 1 shows the set of features incorporated into CredSaT framework. The reader can refer to [60] to obtain further detailed explanations of the methodology used for measuring users' credibility.

*Table 1: Selected features of CredSaT Framework*

| Feature | Description | Equation |
|---|---|---|
| Tweet Similarity Penalty ($Twt\_Sim$) | Represents the count of unique keywords (#*distinctWords*) in the overall user's tweets to the total number of keywords in the user's tweet (#*words*). | $Twt_{Sim_u} = \frac{\#DistinctWords_u}{\#Words_u}$ |
| URL Similarity Penalty ($URL\_Sim$) | Represents the percentage of non-redundant URLs (#*DistinctURLs*) with non-redundant hosts of URLs (#*DistinctURLsHosts*) to the total number of URLs (#*URLs*) posted by user $u$. | $URL_{Sim_u} = 0.5 \times \left(\frac{\#DistinctURLs_u + \#DistinctURLsHosts_u}{\#URLs_u}\right)$ |
| Domain-based content user score ($Sum\_cnt\_scr$) | Is computed by adding all scores retrieved from IBM Watson of tweets' texts posted by user $u$ in domain $d$. | - |
| Domain-based user URL scores ($Sum\_url\_scr$) | Is calculated by accumulating scores for all websites' content of the URLs embedded in user $u$'s tweets in domain $d$. | - |
| Domain-based user scores ($Sum\_all\_scr$) | Refinement summing of the corresponding scores achieved by IBM Watson for all tweets' texts ($Sc_{u,d}^{Twt}$), and the refinement summing of scores retrieved from URLs' webpage content ($Sc_{u,d}^{URL}$) posted by a user $u$ where a domain $d$ was inferred | $Sc_{u,d} = (Twt\_Sim_u \times Sum\_cnt\_scr_{u,d}^{Twt} + URL\_Sim_u \times Sum\_url\_scr_{u,d}^{URL})$ |
| Domain frequency ($DF$) | Count of domains the user $u$ has tweeted about. | - |
| Inverse domain frequency ($IDF$) | Distinguishes users among the list of their domains of interest. | $IDF_u = log(\frac{n}{DF_u})$ |
| Weight ($W$) | Users weights in each domain. | $W_{u,d} = Sc_{u,d} \times IDF_u$ |

| Feature | Description | Equation |
|---|---|---|
| Domain-based user's retweets ($R$) | Represents the frequency of retweets for user' content in each domain $d$ | - |
| Domain-based user's likes ($L$) | Represents the percentage of likes/Favourites count for the users' content in each domain $d$ | - |
| Domain-based user's replies ($P$) | Embodies the count of replies to the users' content in each domain $d$ | - |
| Domain-based user positive sentiment replies ($SP$) | Refers to the sum of the positive scores of all replies to a user $u$ in domain $d$. Positive scores are achieved from IBM Watson with values greater than "0" and less than or equal to "1". The higher the positive score, the greater is the positive attitude the repliers have to the users' content. | - |
| Domain-based user negative sentiment replies ($SN$) | Represents the sum of the negative scores of all replies to a user u in domain d. Negative scores are those values greater than or equal to "-1" and less than "0". The lower the negative score, the greater is the negative attitude the repliers have to the users' content. | - |
| domain-based user sentiments replies ($S$) | Embodies the difference between the positive and negative sentiments of all replies to user $u$ in the domain $d$. | $S_{u,d} = SP_{u,d} - |SN_{u,d}|$ |
| Users' followers ($FOL$) | Total count of users' followers. | - |
| User's friends $FRD$ | Total count of user's friends (followees) | - |
| Followers-friends ratio. $FF\_R$ | User followers-friends ratio. | $FF\_R_u = \begin{cases} \frac{FOL_u - FRD_u}{Age_u}, & \text{if } FOL_u - FRD_u \neq 0 \\ \frac{1}{Age_u}, & \text{if } FOL_u - FRD_u = 0 \end{cases}$, where $Age$ is the age of user profile in years |

As an example of the domain-based credibility analysis, Figure 4 and Table 2 illustrate the key attributes used in the process of conducting credibility analysis on the social data and metadata collected for a well-known politician "Joanne Ryan@*JoanneRyanLalor*" as well as a social spammer "Ham – Hamjuku@*hamjuku*". Figure 4 illustrates the obtained values for certain domain-dependent attributes explained in Table 1. These values are computed for each of the 23 domains inferred from the domain discovery approach that is carried out utilizing IBM Watson API. The values depicted in Figure 4 – (A) demonstrate the domain-dependent analysis of Joanne's tweets which depicts a clear interest in the political domain of knowledge. This is evident considering that she is a member of the Australian House of Representatives and being active in this domain for several years[14]. Figure 4 – (A) also depicts that Joanne's tweets have had quite commended attention from her follower. This can be perceived due to the high number of domain-based likes, retweets, and replies. On the other hand, Figure 4 – (B) shows the domain-based credibility analysis to a social spammer who demonstrated an interest in all domains. This commonly conveys a suspicious behaviour due to the following facts: (i) No one person is an expert in all domains [64]; (ii) A user who posts in all domains does not convey to other users which domain(s) s/he is interested in. A user shows to other users which domain s/he is interested in by posting a wide range of contents in that particular domain; (iii) There is the possibility that this user is a spammer due to the behaviour of spammers posting tweets about multiple topics [65]. This could end up by tweets being posted in all domains which do not reflect a legitimate user's behaviour as in the case of @*hamjuku*.

---

[14] https://en.wikipedia.org/wiki/Joanne_Ryan_(politician), accessed 24-03-2020

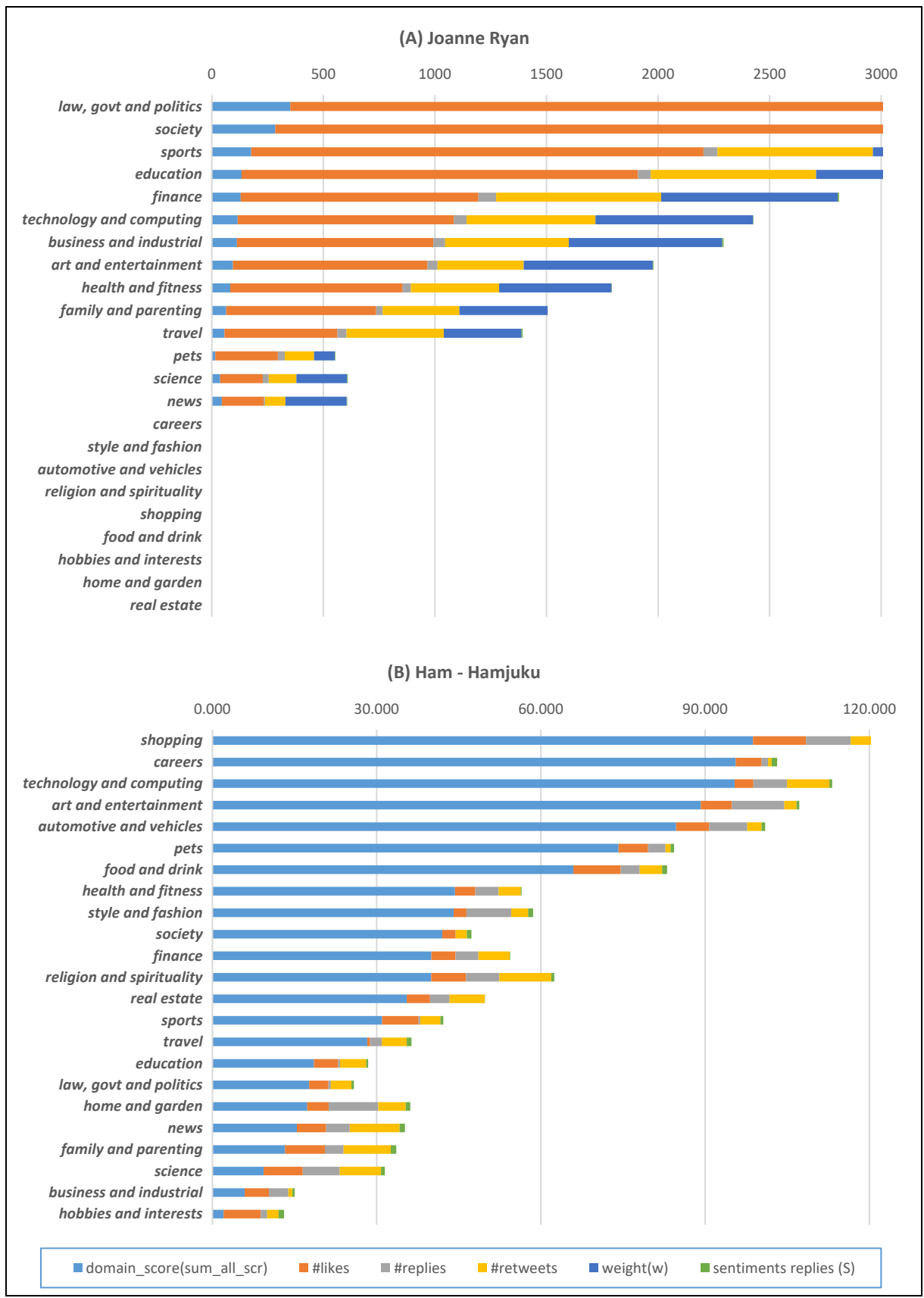

*Figure 4: Domain-dependent social data analysis of two twitterer: (a) Joanne Ryan (MPJoanneRyanLalor); a legitimate politician who is a member of Australian house of representatives, (b) Ham – Hamjuku (hamjuku) : a social twitterer spammer*

Further, Table 2 shows the domain-independent analysis to @*JoanneRyanLalor* and @*hamjuku* twitter profiles. The figures exemplified in this table are plausibly acceptable; the number of users following @*JoanneRyanLalor's* tweets is four times the total number of her friends (i.e. users who follow her). Also, Tweets and URL similarities computed for her 6,495 tweets are around 20% which is quite reasonable. Differently, the similarity analysis computed for both tweets and URLs *of @hamjuku* poses a question on the quality of posted contents; publishing the same content repeatedly is obviously a spammer behaviour [66]. More than 50% of the tweets posted by @*hamjuku* are mainly repeated content. This implies to the textual contents as well as the embedded URLs. The TFF ratio that is calculated for @*hamjuju* sounds rational and legitimate considering the fact that the increase in a number of friends that a user $u$ follows compared to the steadiness in the number of followers commonly indicates a suspicious behaviour, and such a user is likely to be a spammer [65, 67]. However, as it can be inferred from the analysis conducted to @*hamjuku*, friends to followers ratio analysis could not be considered as sole spamming detection criterion, and this does not necessarily exhibit a credible profile; further scrutiny should be carried out to examine the overarching behaviour of a spammer, thereby providing a reliable detection mechanism.

*Table 2: Domain-independent social data analysis for a legitimate twitterer and a spammer twitterer*

| Feature | Joanne Ryan | Ham – Hamjuku |
|---|---:|---:|
| #followers | 5,606 | 248 |
| #friends | 1437 | 120 |
| Age | 7 | 13 |
| **TFFRatio** | **595.571** | **9.846** |
| #Tweets | 6,459 | 3,893 |
| #DistinctWords | 24,795 | 5,392 |
| #Words | 112,889 | 10,733 |
| **Twt_Sim** | **0.212** | **0.502** |
| #DistinctURLsHosts | 85 | 5 |
| #DistinctURLs | 291 | 150 |
| #URLs | 861 | 5,591 |
| **URL_Sim** | **0.218** | **0.528** |

### 3.5 Knowledge Graph Creation

At this stage, the knowledge representing the politics domain and the incorporated credible users and their data and metadata are captured in the domain ontology. In addition, knowledge is depicted in a less expressive relational model that stores knowledge obtained from the analysis conducted on users' social metadata and inferred from their collected textual content. The relational model embodies also the users' domain-based credibility indicating the trustworthiness of the users in each domain of knowledge. Knowledge graph creation module aims to transform the collected heterogeneous data format onto a unified standard form.

The Resource Description Framework (RDF) is a widely used underlying model to represent knowledge in terms of triples (*subject, predicate, object*), where the subject of the triple indicates the resource which needs to be described, predicate indicates the property of the subject, and object refers to the property value which describes the subject. A typical knowledge graph is represented as a directed graph where nodes indicate the entities (resources) of the class model and edges depicts the relations (properties) between those entities.

The datasets collected in this study are in different formats; Tabular, JSON, and CSV. One of the crucial steps in conducting Big data analytics is to provide a consolidated platform to handle the heterogeneity of the datasets collected from diverse data islands. Hence, we incorporate The RDF Mapping Language

(RML)[68] as a mapping language to express data in dissimilar format into a unified RDF form, thereby mitigating the variety dimension of Big data[69]. RML defines a generic approach for mapping different data structures, where the input could be any data source and the provided output is provided as an RDF graph. The mapping process in RML consists of one or more *triple maps*. Each triple map embodies a logical source (input source), subject map (describes the mechanism to generate the subject for each logical resource), and predicate-object-map (specifies the predicate and the object map and how the triple's predicate is generated). RML mapping rules are used in this study to transform the annotated components into RDF triples to enrich the knowledge graph of the semantic repository.

For annotation and enrichment, the domain knowledge graph is fed with annotated politics entities extracted from the textual contents of the tweets. The annotation is then enriched with a description of the concepts referring to the domain ontologies and using controlled vocabularies e.g. Dublin Core (DC[15]), Simple Knowledge Organization System (SKOS[16]), Semantically-Interlinked Online Communities (SIOC[17]). This allows each entity in the textual data to be specified with its semantic concept. The particular concepts can be further expanded into other related concepts and other entities instantiated by the concepts. The consolidation of this semantic information provides a detailed view of the entities captured in domain ontologies.

For the interlinking process, entities are interlinked with similar entities defined in other datasets to provide an extended view of the entities represented by the concepts. Our focus is on equivalence links specifying URIs (Universal Resource Identifiers) that refer to the same resource or entity. Ontology Web Language (OWL) provides support for equivalence links between ontology components and data. The resources and entities are linked through the 'owl#sameAs' relation; this implies that the subject URI and object URI resources are the same. Hence, the data can be explored in further detail. In the interlinking process, different vocabularies i.e. Upper Mapping and Binding Exchange Layer (UMBEL), Freebase – a community-curated database of well-known people, places, and things, YAGO – a high-quality knowledge base, Friend-of-a-Friend (FOAF ), Dublin Core (DC ), Simple Knowledge Organization System (SKOS ), Semantically-Interlinked Online Communities (SIOC ), and Google KG, are used to link and enrich the semantic description of resources annotated.

The domain KG is also enriched with knowledge inferred from the social presence of users on twitter platform. This primarily encompasses associated metadata of the users and their content, such as #followers, #friends, #likes/favourites, and #retweet/share, etc. It also includes the resultants values obtained from the conducted domain-based credibility analysis to users. This includes values of the number of domains the users are interested in, the credibility value of the user in each domain of knowledge, the number of political entities indicated in the user's tweet, the number of the positive, negative and neutral replies to the user's tweets, etc.

Figure 5 illustrates an example of an RDF graph of knowledge inferred from multi-resources heterogeneous data collected for the politician, Joanne Ryan. This RDF graph can be referred to an RDF molecule as it represents a set of RDF triples indicating the same subject. This RDF molecule has been constructed as a result of the transformation process conducted on the data by the means of defined rules of RML mapping. The RML mapping rules are used further to ensure the format of the designated unique identifiers (URIs) for the mapped resources which are used as the subject of all the RDF triples.

---

[15] https://www.dublincore.org/
[16] http://www.w3.org/2004/02/skos/
[17] http://sioc-project.org/

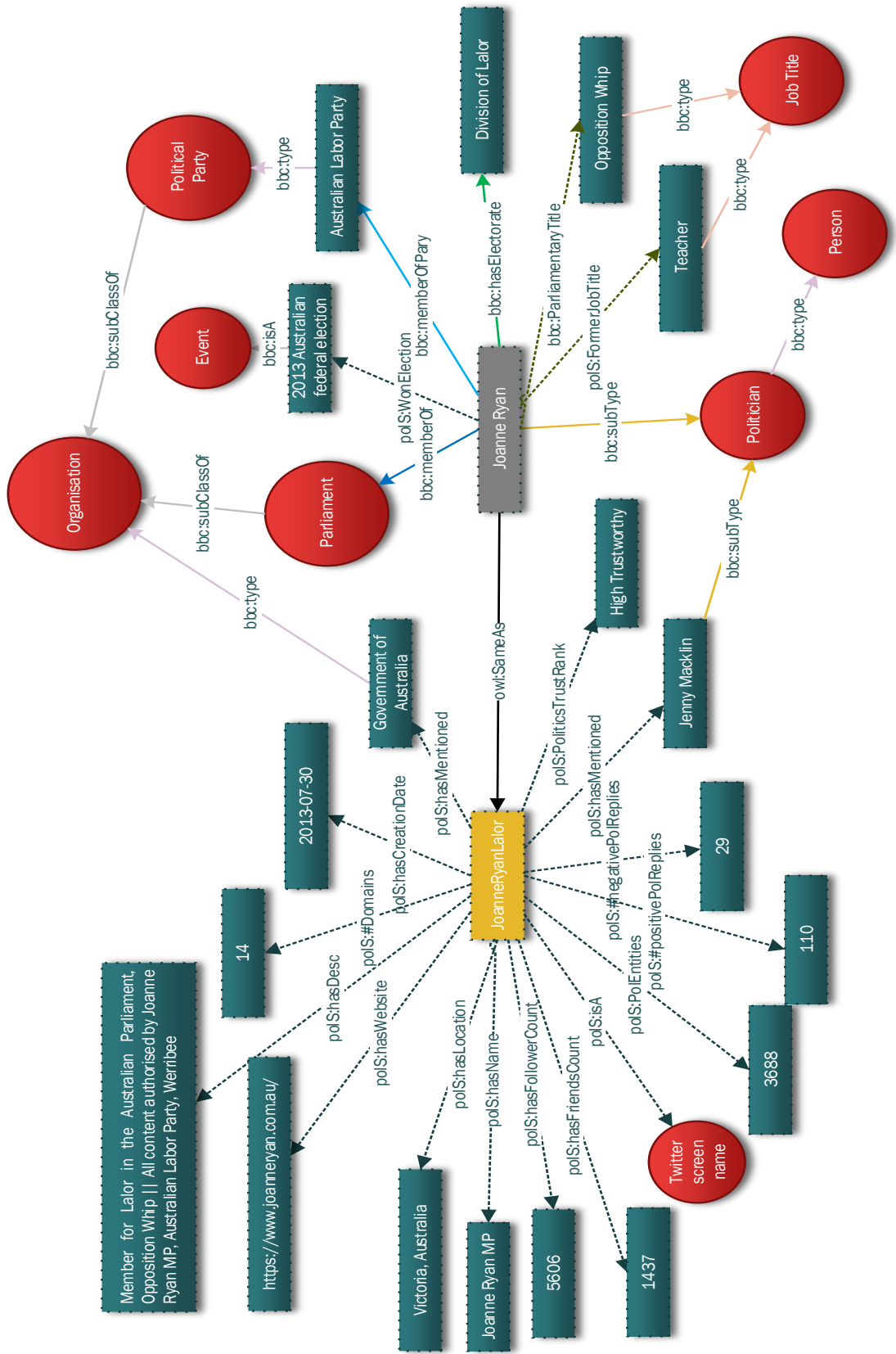

Figure 5: Example of an RDF molecule describing "Joanne Ryan" obtained from the KG Creation process.

## 3.6 Knowledge Graph Embedding Models

Knowledge Graph Embedding (KGE) is the process of transforming the constituents of a KG (entities and relationships) into a low-dimension semantically-continuous space [18]. Despite the fact that solving problems pertaining to graphs can be carried out on the conventional graph presentation (i.e. adjacency matrix), mapping the entire graph or its nodes to the vector space has attracted the scientific community due to its scalability to simplify resolving several complex real-life graph problems such as KG completion, entity resolution, and link-based clustering, just to cite a few [41, 42, 70]. Embedding a KG is learned via training a neural architecture over a graph, and comprises commonly three main components, namely; (i) encoding entities into distributed points in the vector space, and encoding relations as vectors, or other forms; (ii) scoring function or model-specific function that is used to evaluate the model's efficiency; (iii) optimization procedure, which aims to learn the optimal embedding for the designated KG, thereby the scoring function assigns high scores to positive statements.

The literature in KG Embedding commonly categorises the embedding techniques into two main classes; translation distance models and semantic matching models [18]. Translation Distance Models are designed to evaluate the plausibility of a certain fact in a distance between two entities. Semantic Matching Models intend to measure the plausibility of facts considering the latent semantics of entities and relations into their low dimensional representations. Amongst numerous KG embedding models proposed in the literature, the following are the set of most popular KG embedding models that are incorporated in this study.

**Translating Embedding (TransE)[71]:** learns the representation of both the entities and relations as vectors in the same low dimensional semantic space. Hence, for a golden triple $(h, r, t)$, **TransE** treats the relation $r$ as a translation in the embedding space so that $h + r \approx t$, when $(h, r, t)$ holds ($t$ should be the closest to $h + r$), otherwise $h + r$ should be away in distance from $t$. The distance between the entities after conducting transformation measure the accuracy of the relation by using $L1$ or $L2$ norm. $L1$ or $L2$ norm($||\cdot||$) are used with the scoring function to compute the similarity between the embedding of the head that is translated by the embedding of the relation, and the embedding of the tail as follows:

$$f_{TransE} = -||e_h + e_r - e_t||_n \quad (1)$$

The scoring function is expected to be large if triple $(h, r, t)$ holds. The scoring function then can be used on positive and negative triples $t^+, t^-$ in the loss function.

**The DistMult model[72]:** is an extension and a simplification to RESCAL[73] and is based on the bilinear model. In this model the relation is encoded as diagonal (single vector) using the trilinear dot product as scoring function:

$$f_{DistMult} = r^T(h \odot t) \quad (2)$$

**Complex Embeddings (ComplEx)[74]:** this is an extension to DistMult model by introducing complex-valued embeddings, where the scoring function is based on the trilinear Hermitian dot product in $C$:

$$f_{ComplEx} = Re(\langle e_r, e_h, \bar{e_t} \rangle) \quad (3)$$

Entity and relation embeddings are no longer positioned in real space but in a complex space.

**Holographic Embeddings(HolE)[43]**: a compositional vector space model that learns compositional vector space representations of entities and relations through incorporating the strength of RESCAL as well as the simplicity of **DistMult.** The scoring function of **HolE** is defined as:

$$f_{\text{HolE}} = w_r \cdot (e_h \otimes e_t) = \frac{1}{k} F(w_r) \cdot (\overline{F(e_h)} \odot F(e_t)) \qquad (4)$$

HolE is based on the circular correlation(denoted by $\otimes$) of vectors that makes compression of pairwise interactions and that can capture rich interactions in relational data.

**Convolutional 2D KG Embeddings (ConvE)[75]:** is a neural link prediction model that uses deep, multi-layer, conventional and fully connected layers of nonlinear features to tackle the interactions between input entities and relations:

$$f_{\text{ConvE}} = \langle \sigma(vec(g([\overline{e_h}; \overline{e_r}] * \Omega))W))e_t \rangle \qquad (5)$$

where $g$ is a non-linear activation function, $W$ is the convolutional filters, $*$ is the linear convolution operator, $vec$ vectorization operation indicates 2D reshaping.

**Convolution-based model (ConvKB)[76]:** incorporates conventional neural networks to represent the concatenation of entities and relations, which increases the learning ability of latent features. The scoring function of **ConvKB** is defined as follows:

$$f_{\text{ConvKB}} = concat(g([e_h, e_r, e_t] * \Omega)) \cdot W \qquad (6)$$

where $g$ is a non-linear activation function, $concat$ is the concatenation operator, $W$ is the convolutional filters, $*$ is the linear convolution operator, $\cdot$ is the dot product and $\Omega$ is a set of filters.

## 3.7 Knowledge Graph Embedding Model Evaluation

In this section we provide a brief on the performance metrics incorporated in this study to measure the performance of the KG Embedding models as well as on link prediction task.

The following learning-to-rank metrics are commonly used in the literature to measure the performance of KG Embedding models:

- Mean Reciprocal Rank (MRR): is a function that computes the mean of the reciprocal of elements embodied in a vector of rankings. It is used as a measure to evaluate the system performance against the retrieved elements. The formal definition of MRR is:

$$MRR = \frac{1}{|T|} \sum_{i=1}^{|Q|} \frac{1}{rank(s,p,o)_i} \qquad (7)$$

where $rank(s, p, o)_i$ refers to the rank of a positive element $i$ against a list of negative elements, $T$ is a set of test triples and $(s, p, o)$ is a triple $\in T$.

- Mean Rank (MR): refers to the mean rank of the correct test facts/triples embodied in a vector of rankings (i.e. the average of the predicted ranks). MR can be computer as follows:

$$MR = \frac{1}{|T|} \sum_{i=1}^{|Q|} rank(s,p,o)_i \qquad (8)$$

- $Hits@N$: indicates the number of elements in the ranking vector retrieved from the model are positioned in the top ($N$) locations. $Hits@N$ can be defined as follows:

$$Hits@N = \begin{cases} \sum_{i=1}^{|Q|} 1 & , \text{ if } rank(s,p,o)_i \leq N \\ 0, & otherwise \end{cases} \qquad (9)$$

$Hits@N$ provides the probability that the correct results appear on the top where the proportion of the ranks does not exceed $N$. When $N = 1$ it checks if the target test fact can be correctly predicted from the first attempt. Embedding models Commonly achieve better results when the value of ***N*** is higher (for example 10).

Another key metric is used to evaluate the performance of link prediction task. The metric is composed of the three well-known measures, namely: *Precision*, *Recall* and *F-Score*, that are commonly used in classification tasks. Further discussion on these metrics and their usage is depicted in Section 4.4.

## 4. Experimental Results

### 4.1 Dataset Selection

As indicated previously this study aims to construct a domain-based KG (politics) and to carry out embeddings on the constructed KG that will assist in conducting further analysis. We make use of the twitter platform to consolidate the domain-based KG with facts inferred from social contents propagated from this virtual platform. As indicated in section (3.2.1 Dataset Acquisition), the dataset is collected from dissimilar resources based on three different categories of users: (A) Members of the Australian house of presentative (Senators and MPs); (B) users interested in Politics domain; and (C) users whose domain of interest is not explicitly conveyed. This set also might contain spammers, anomalous users, and other untrustworthy users. Hence, amongst 253 of the selected users of Category (C), 40 users are detected as spammers, hence their data are eliminated from further analysis. Table 3 shows some figures on the collected datasets for each category.

*Table 3: Collected datasets*

| Users category | #users | #tweets | #entities | #political_entities | #facts(triples) |
|---|---|---|---|---|---|
| (A) Politicians (senators and members of parliament) | 227 | 611,739 | 1,461,645 | 172,095 | 16,286 |
| (B) Politics-interested | 622 | 1,540,723 | 1,815,091 | 136,458 | 34,149 |
| (C) Unknown politics interest | 253 (40 filtered) | 221,840 | 845,293 | 6,633 | 3,062 |

Domain analysis using IBM Watson has been conducted on each category of users to infer the domains of interest for each category. Figure 6 illustrates the total number of users and their tweets distributed over 23 domains of knowledge for each designated category. As depicted in Figure 6, category (A) shows a clear interest in the politics domain which is reasonable considering this category of users are mainly politicians, and their social contents are expected to discuss topics related to politics. Category (B) is a mixture of users who are selected as they explicitly show a common interest in politics. The domain analysis on category (B) supports this and shows that those users are interested in politics as well as other domains such as technology, art and entertainment, and travel. Despite the slight interest in politics domain and a strong interest in other venues, Category (C) demonstrates a balanced interest across the topics of interest.

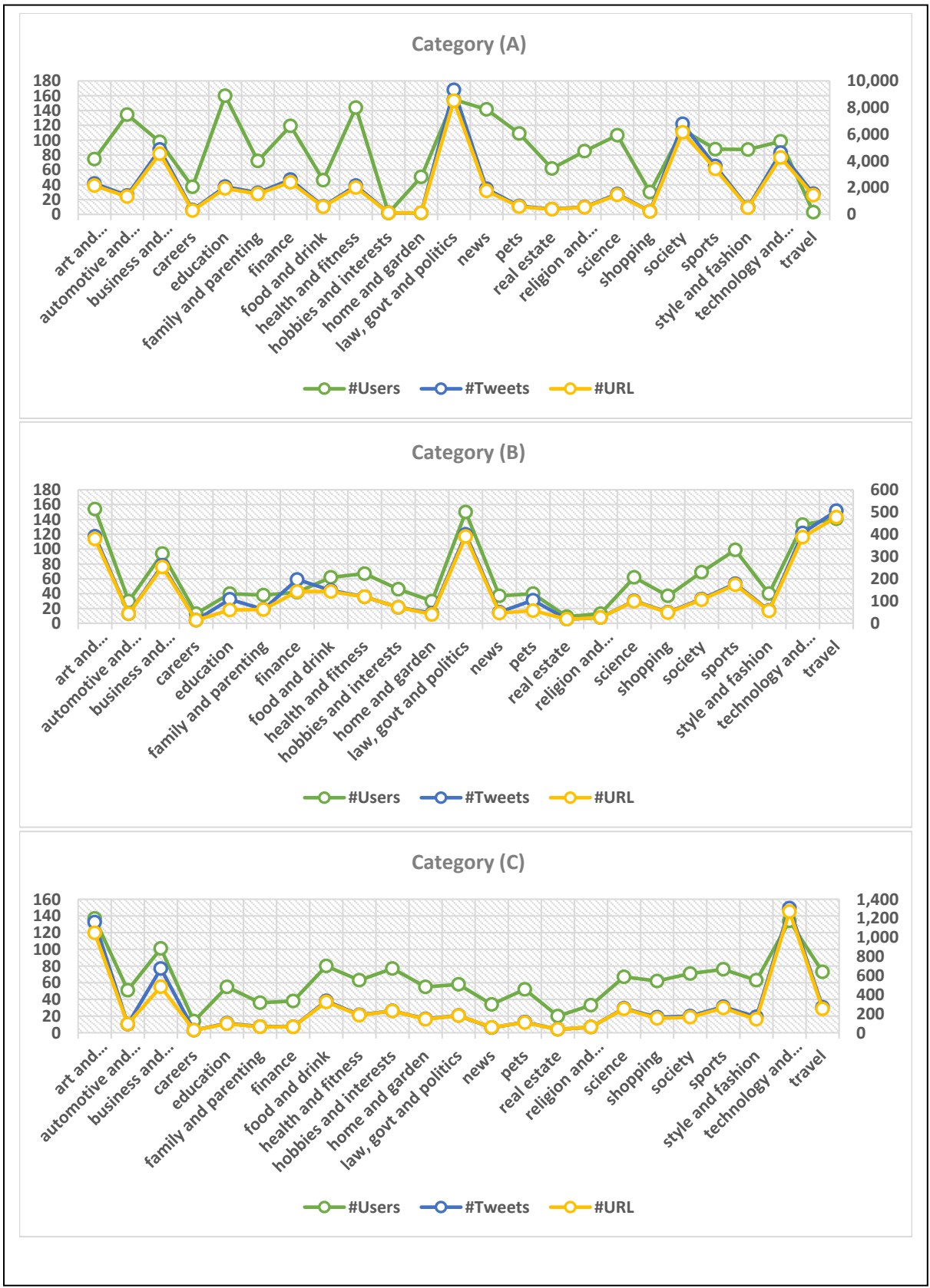

*Figure 6: The distribution of the total number of users and posted tweets in each designated domain for three categories: (A) Politicians (senators and members of parliament), (B) Politics-interested and (C) Unknown politics interest.*

## 4.3 Domain KG Embedding Model Evaluation

### 4.3.1 Experiment Settings

This study incorporates Ampligraph™ [77] version 1.3.1, with TensorFlow 1.14, and Python 3.7 on the backend for conducting KG embeddings on the constructed domain KG. All the experiments including training and evaluation of each embedding model were carried out using the Australian Pawsey supercomputing high-performance facilities[18]. The domain KG is initially divided onto training, test, and validation subsets. Several KG embedding models are implemented and their hyperparameters are tuned using the random search strategy. Random search has proven efficiency and outperformed grid search routine as it provides a solid baseline, and it also shows robustness when the number of parameters increases [78, 79]. A brief description of some internal settings used in the incorporated embedding models is provided in Table 4.

*Table 4: Hyperparameters used for the KG Embedding models*

| Hyperparameter | Description |
| --- | --- |
| Batches_count | #batches to complete one epoch of the Platt scaling training |
| seed | The value of the seed used to infer random numbers |
| epochs | #iterations used to train the Platt scaling model |
| k | Embedding space dimensionality |
| eta | #Negatives that must be generated at runtime during training for each positive |
| loss | Loss function used to train the model. Examples: { pairwise, nll, absolute_margin, self_adversarial, multiclass_nll } |
| loss_params | Set of parameters used for loss-specific hyperparameters |
| regularizer | The strategy used with the loss function |
| regularizer_params | Set of parameters for regularizer-specific hyperparameters |
| optimizer | The optimizer used to update the weight parameters thereby minimizing the loss function. Examples: {'sgd', 'adagrad', 'adam', 'momentum'}. |
| optimizer_params | Arguments passed as parameters to the optimizer, such as learning rate(lr) and momentum. |
| verbosebool | The verbose mode |

Table 5 shows the set of hyperparameters tested using the random search strategy, and those underlined are the optimal values obtained from the incorporated search strategy.

*Table 5: The Embedding model and the incorporated hyperparameters used in the random search*

| Embedding Model | Examined Hyperparameters (optimal settings obtained from the random search are in bold format) |
| --- | --- |
| TransE | *batches_count*= {50, **100**, 150}, *seed*= {0, **555**}, *epochs*= {500, 1000, **2000**, 4000}, *k*= {**100**, 200}, *eta*= {**5**, 10, 15, 20}, *optimizer*= {'adam', **'adagrad'**}, *loss*={**'pairwise'**,'nll', 'absolute_margin'}, *verbose*= {True, **False**}, *regularizer*= {None, **'LP'**} |
| DistMult | *batches_count*= {**50**, 100, 150}, *seed*= {**0**, 555}, *epochs*= {500, 1000, **4000**}, k= {**100**, 200}, *eta*= {5, 10, **15**, 20}, *optimizer*= {'adam', **'adagrad'**}, *loss*={**'pairwise'**, 'nll', 'absolute_margin' }, *verbose*= {True, **False**}, *regularizer*= {**None**, 'LP'}, *normalize_ent_emb* = {**True**, False} |

---
[18] https://pawsey.org.au/

| Embedding Model | Examined Hyperparameters<br>*(optimal settings obtained from the random search are in bold format)* |
|---|---|
| ComplEx | *batches_count*= {50, 100, 150, **200**}, *seed*= {**0**, 555}, *epochs*= {500, **1000**, 4000}, k= {100, **200**}, eta= {5, 10, **15**, 20}, *optimizer*= {'adam', **'adagrad'**}, *loss*={**'pairwise'**,'nll', 'absolute_margin'}, *verbose*= {True, **False**}, *regularizer*= {**None**, 'LP'} |
| HolE | batches_coun= {50, **100**, 150}, seed= {**0**, 555}, epochs= {500, 1000, **4000**}, k= {100, **200**}, eta= {**5**, 10, 15, 20}, optimizer= {'adam', **'adagrad'**}, loss={'pairwise', **'nll'**, 'absolute_margin' }, verbose= {**True**, False}, regularizer= {None, **'LP'**} |
| ConvE | batches_count= {50, **100**, 150}, seed= {0, **555**}, epochs= {500, **1000**, 4000}, k= {100, 200}, eta= {5, 10, **15**, 20}, optimizer= {**'adam'**, 'adagrad'}, loss={**'BCE'**}, verbose= {True, **False**}, regularizer= {None, **'LP'**}, conv_filters= {24, **32**}, conv_kernel_size = {1, **2**, 3}, dropout_embed= {0,2,0.3}, dropout_conv= {**0.2**, 0.3}, dropout_dense= {01,**02**} |
| ConvKB | batches_count= {50, 100, **150**}, seed= {0, **555**}, epochs= {500, 1000, **4000**}, k= {100, **200**}, eta= {5, **10**, 15, 20}, optimizer= {'adam', **'adagrad'**}, loss={**'pairwise'**,'nll', 'absolute_margin' }, verbose= {**True**, False}, regularizer= {None, **'LP'**}, num_filters= {**24**, 32}, filter_sizes= {1,**2**,3}, dropout = {0.0,**0.1**} |

### 4.3.2 Evaluation Protocol

This study incorporates the evaluation protocol proposed by [71]. There are three key steps in the defined protocol, namely: (i) generating negative triples synthetically; (ii) remove the resultant positive triplets; and then (iii) ranking all the test facts (triples) against the triples returned from the preceding step. Negative triples are initially positive triples (correct facts) which have been manipulated (corrupted) by randomly replacing head, tail or relation, thus creating new triples(false facts). The negative sampling mechanism used in this paper is based on corrupting the head and tail one side each time, to be acquiescent with the local closed-world assumption indicated in [43], then we compute the average of attained evaluation metrics of each method.

### 4.3.3 Embedding Evaluation Results

The experiments have been carried out incorporating six well-known embedding models as depicted in Table 4 along with the depicted tuned hyperparameters. With the ranks obtained from the subjects and predicates corruption described in the previous subsection, the metrics are computed for each embedding model. Figure 7 illustrates the attained metric values obtained from each model. Despite the convergence in the outcome performance results, ConvKB embedding model outperforms other models in all key metrics. For example, examining $hits@1, hits@3, and\ hits@10,$ with ConvKB, we were able to hit a correct subject or predicate 55%, 66.9%, and 81% of the times respectively. This interpretation applies to all other metric values obtained from each embedding model. The good performance of ConvKB is established due to the underlying structure of ConvKB; it incorporates a CNN network to capture the global relationships and the transitional features of the KG embodied entities and relations. HolE model has also shown promising results; this is understandable as Hole integrates the efficiency and simplicity of more than one model [18]. Moreover, Hole can obtain rich interactions in such relational data by applying circular correlation on vectors that create compositional representations.

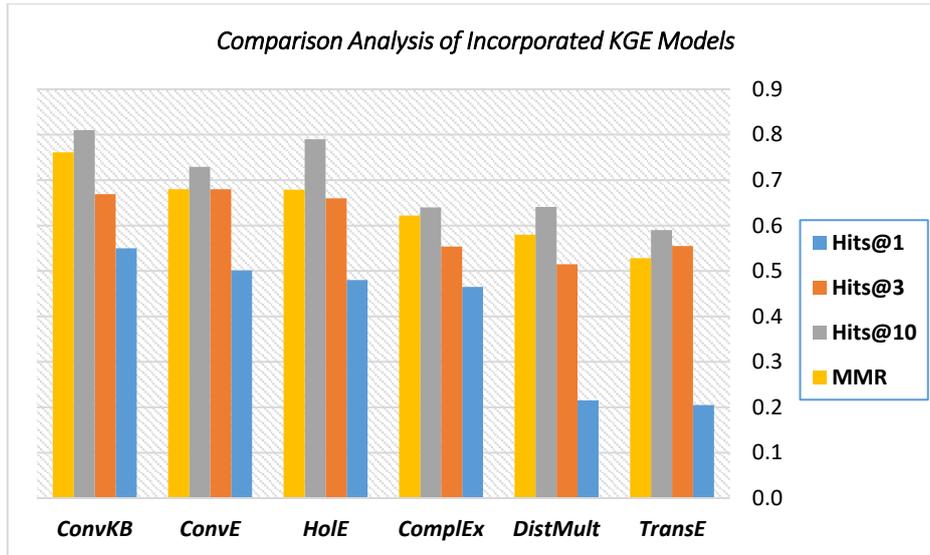

*Figure 7: Comparison analysis of evaluation metrics of the incorporated embedding models*

The utility of the embedding models commonly measured by the applicability of using these models in more factual tasks. The following sections discuss the utility of the developed approach in link prediction, clustering, and visualisation tasks.

### 4.4 Task (1): Link Prediction

The implemented KG Embeddings in this study are used to carry out a Link Prediction task. We have generated a set of facts in politics domain, which contain true political facts that have not been trained in the model (unseen facts) as well as some synthetically created false politics facts. The goal is to test the utility of the model to detect which of the presented true candidate facts are likely to be true. Similarly, which false candidate facts are unlikely to be true.

As depicted in Figure 8, the applicability of each embedding model in a prediction task can be also evaluated by determining, whether or not a statement's label is true. The filled diamonds are classified as true politics facts, and the non-filled diamonds are false politics facts. Four dissimilar states can be obtained from this diagram as follows:

- True-Positives (TPos): refer to the true politics facts that are classified by the model as true politics statements.
- False-Positives (FPos): refer to the false politics facts that are classified incorrectly as true politics statements.
- True-Negatives (TNeg): refer to the false politics facts that are classified correctly as false politics statements.
- False-Negatives (FNeg): refer to the true politics facts that are classified incorrectly as false politics statements.

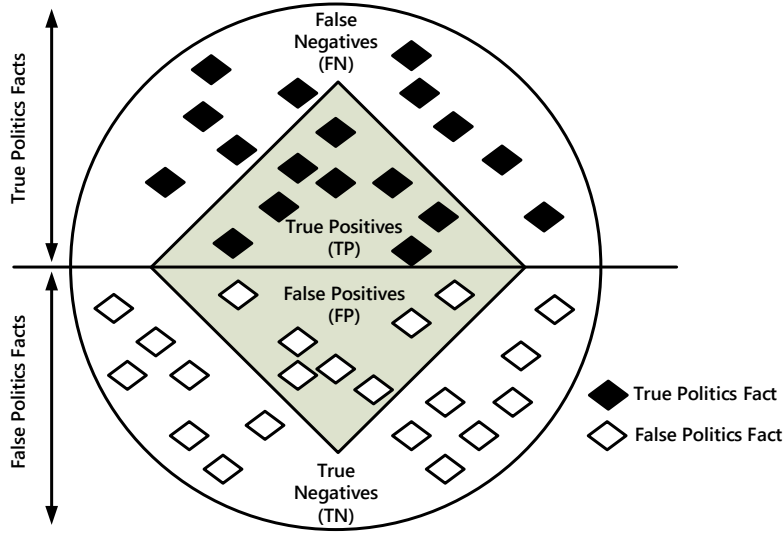

*Figure 8: Classification of True politics facts and False politics facts*

Table 6 shows the confusion matrix which gives a further depiction of the performance of the four aforementioned states which will be used to evaluate the performance of the prediction module.

*Table 6: Confusion matrix*

|  |  | Predicted Facts | |
|---|---|---|---|
|  |  | **True Fact** | **False Fact** |
| **Actual Facts** | **True Fact** | *TPos* | *FNeg* |
|  | **False Fact** | *FPos* | *TNeg* |

The assessment to the prediction task has been achieved by incorporating the following metrics to evaluate the utility of the embedding model in predicting whether or not a certain political fact is true:

**Accuracy:** this metric specifies the accuracy of the incorporated embedding model in making a correct prediction. Accuracy is the ratio obtained between the accurate predictions (i.e. $TPos + TNeg$) and the overall inferred predictions ($FNeg + TPos + FPos + TNeg$). Accuracy can be computed as:

$$Accuracy = \frac{TPos+TNeg}{FNeg+TPos+FPod+TNeg} \qquad (10)$$

**Precision**: this metric refers to the proportion of those facts that were classified accurately as positive and they are actually positive. It can be calculated as:

$$Precision = \frac{TPos}{TPos+FPos} \qquad (11)$$

**Recall**: recall indicates the proportion of true positive facts were correctly classified as positives facts. It can be computed as:

$$Recall = \frac{TPos}{TPos+FNeg} \qquad (12)$$

**F- measure**: f- measure (f-score) is a harmonic measure used to provide a trade-off between precision and recall. It can be computed as:

$$F-score = \frac{Precision \cdot Recall}{Precision+Recall} \quad (13)$$

Precision specifies the proportion between the sum of actual true politics facts that are accurately predicted and the total sum of accurate and inaccurate predictions of true politics facts. Recall specifies the proportion between the number of actual true politics facts that are accurately predicted and the total sum of actual true politics facts. Therefore, obtaining high precision value indicates that the prediction module is a success in the result relevancy measure and is able to deduce more relevant politics facts among the retrieved ones. Attaining high recall value indicates that the prediction module is a success in retrieving truly relevant results. For example, if a prediction module is evaluated and attains a precision value of '1', this indeed conveys that all predicted facts are correct predictions and depict factual politics facts that can be used to augment the knowledge graph. However, this does not necessarily reflect the module's efficacy to retrieve all true politics facts. On the other hand, if the prediction module attains a recall value of '1', this implies that this module is able to retrieve all true positive facts. Yet, it does not convey the number of other false retrieved predictions. This is why it is commonly a good practice to incorporate the F-measure metric as it provides a weighted average of precision and recall.

The ground truth dataset of the link prediction experiment contains 1,000 labelled statements of both true politics facts as well as false politics facts. Figure 9 shows the performance comparison on the link prediction task for six incorporated embedding models. As depicted in the diagrams, ConvE embedding model has relatively overshadowed other embedding models in this task. For example, this designated model has obtained 74.4%, 83.2 %, 77.2, 80.1% in accuracy, precision, recall, and f-score metrics respectively, where 51.4% of the true politics facts were correctly predicted as true politics facts and 23% of the false politics facts were correctly classified as false politics facts. HolE and ConvE have also shown promising performance results. For example, ConvE was able to predict almost half of the true positive facts correctly as true positive statement. Also, only 18.3% of false politics facts observed by ConvE were incorrectly assorted as true politics facts, and 21.3% of true politics facts were assorted incorrectly as false politics statements. Therefore, the precision computed for ConvE was 72% which proves the ability of this embedding model to demonstrate virtuous results in this task.

On the other hand, Figure 9 shows that the results of TransE, DistMult and ComplEx embedding models are convergent in almost all computed metrics (i.e. accuracy, precision, recall, and f1-score). In spite of the fact that TransE performs well in datasets that embody one-to-one relationships, it demonstrates inadequacy to handle unbalanced relations (i.e. one-to-many/many-to-one) [14]. For example, embedding two knowledge facts such as; (Anthony Albanese, hasLocation, NSW) and (John Alexander, hasLocation, NSW) will result in "Anthony Albanese" entity vector be close to "John Alexander" entity vector. However, this does not convey the factual and realistic similarity between these two politicians; "Anthony Albanese" is a member of the Australian Labor Party while the affiliation of "John Alexander" is Liberal Party. This discrepancy also applies to their electorate and other facets. Furthermore, DistMult embedding model is inadequate to handle asymmetric and antisymmetric relations. This is evident because of the entry-wise product depicted in Eq. (2); it demonstrates that all relations are symmetric. This attains misleading results when asymmetric and antisymmetric relations are present[80]. HolE, on the other hand, is skillful to address this issue since it uses a circular correlation operator, this results in HolE able to capture the relations with asymmetricity and anti-symmetricity [81].

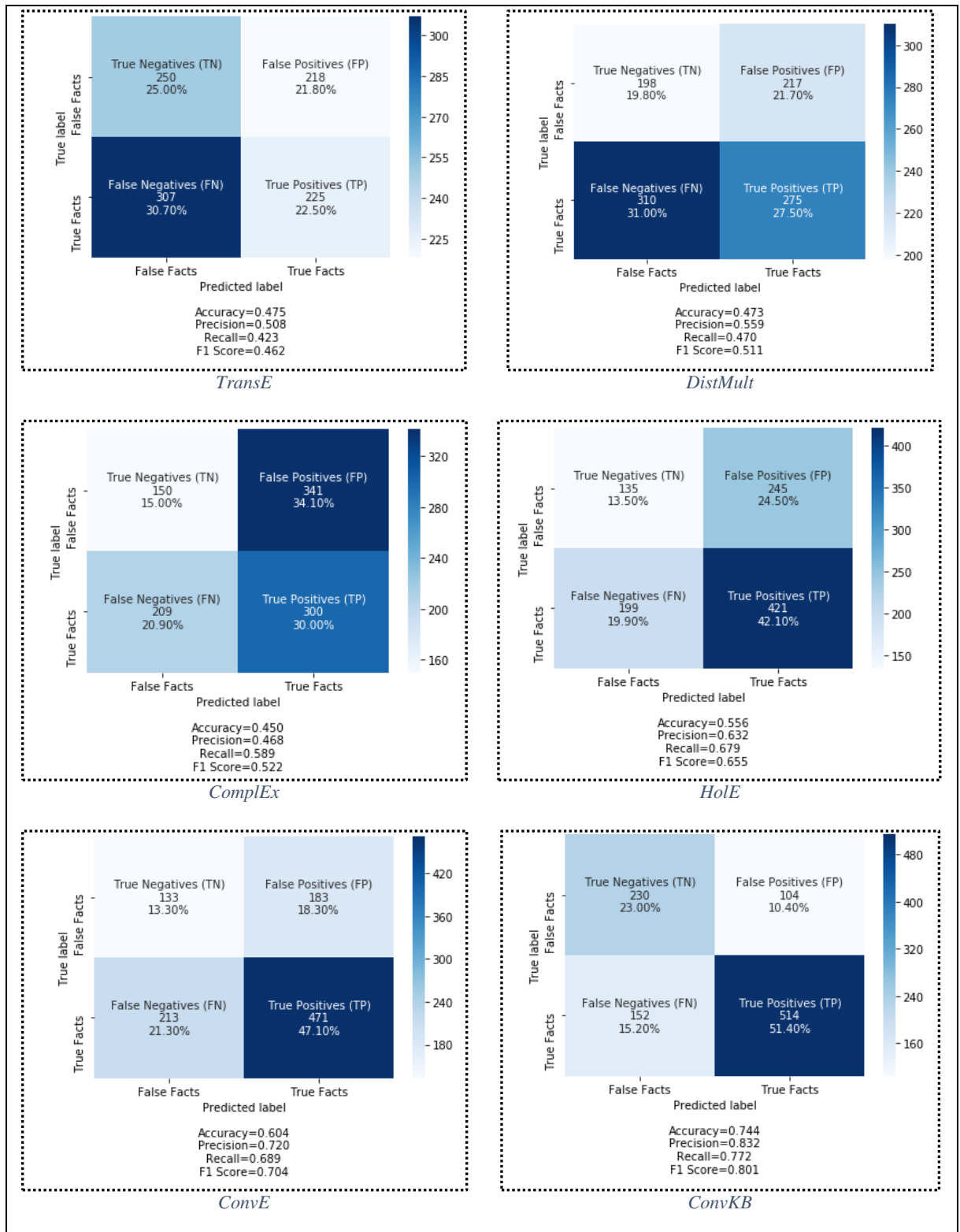

*Figure 9: Performance Comparison on Link Prediction task for six KG embedding models*

Table 7 shows an example of the link prediction task. The table presents a set of selected statements obtained from the ground truth, each statement with a label indicating, whether the statement is true or false along with the classification label acquired from ConvKB embedding model. It can be seen that the embedding model has been largely able to understand Australian politics and provide some good prediction on this domain. For example, the model is able to indicate that Karen Andrews is actually a member of the Australian Labor Party despite the fact that this information was not imported to the KG.

It also able to discover political interest of users whose domain of interest is not explicitly depicted. For example, the collected tweets of @JohnKeily1 demonstrate that this user is interested in politics and does not support the Australian Labor Party (ALP). The model captures some truth about this user and did detect that this user is highly interested in politics, yet it fails to capture that @JohnKeily1 is not a supporter of ALP party. This can be understood considering that @JohnKeily1 has some negative tweets about ALP party and his presence in the vector space turns out to be close to those supporting this political party. This explanation can be also applied to other instances where the model was unable to correctly classified them. Hence, in the future work, the KG will be further scrutinised and enhanced to embody for example the sentiments of the social contents, political polarisation, etc.

*Table 7: A selected set of labelled candidates (true and false facts)*

| Unseen Statement | | | Label | Prediction |
|---|---|---|---|---|
| *Subject* | *Predicate* | *Object* | | |
| Karen Andrews MP | memberOfParty | Australian Labor Party | *True* | *True* |
| jocey70 | memberOfParliament | Australian Parliament | *False* | *True* |
| Nic Hodges | hasSubtype | Politician | *False* | *True* |
| JohnKeily1 | hasPoliticsInterest | High | *True* | *True* |
| Jay McCormack | hasMentioned | Australian Parliament | *False* | *False* |
| Karen Andrews MP | memberOfParliament | Australian Parliament | *True* | *True* |
| Dgoodlad | memberOfParliament | Australian Parliament | *False* | *True* |
| Bridget Archer MP | hasMentioned | Australian Labor Party | *False* | *False* |
| Bridget Archer MP | memberOfParty | Liberal Party of Australia | *True* | *False* |
| Kevin Andrews MP | supports | Liberal Party of Australia | *True* | *True* |
| JohnKeily1 | supports | Australian Labor Party | *False* | *True* |
| stevene | memberOfParty | Australian Greens | *True* | *True* |
| Katie Allen MP | memberOfParty | Australian Labor Party | *False* | *False* |

## 4.5 Task (2): KGE Clustering and Visualisation

Cluster analysis is another evaluation strategy that can be performed on a constructed KG. Clustering occurs on embedding space of both entities and relations and it is an effective evaluation strategy to measure the subjective quality of the KG embedding. The clustering projects the original embedding with the predetermined space size into a 2D space, then a subjective measure is carried out to evaluate the embeddings. Several clustering algorithms[19] have been implemented and evaluated, such as *AffinityPropagation, AgglomerativeClustering, Birch, DBSCAN, FeatureAgglomeration and KMeans*. Several projections have been generated from these clustering modules, yet *KMeans* algorithm has proven effective due to the factual projections that are generated by this algorithm. This experiment follows the standard embedding space size (i.e. k = 100) used by AmpliGraph™.

Figure 10 illustrates that four clusters have been generated for a set of selected politicians and users interested in politics. The depicted cluster is widely accepted; it can be seen from the figure that almost all the members who have been categorised to the same group have the same political attachment. For example, Kevin Andrews MP, Lucy Wicks MP, Rowan Ramsey MP to name a few are all members of the Liberal Party of Australia and have been grouped to the same cluster. Likewise, Matt Keogh MP, Fiona Phillips, and Brian Mitchell are also assembled with others in the same cohort (i.e. ALP). Further, the figure depicts that the incorporated clustering approach is also able to infer politics affiliation of non-politicians; for example, the twitterer (@wheels002) happens to appear in the same vector space with members of Australian Parliament who are also members of Australian Greens Party. This is evident since *@wheels002* has conveyed her interest in this political party in several tweets.

---

[19] https://scikit-learn.org/stable/modules/clustering.html

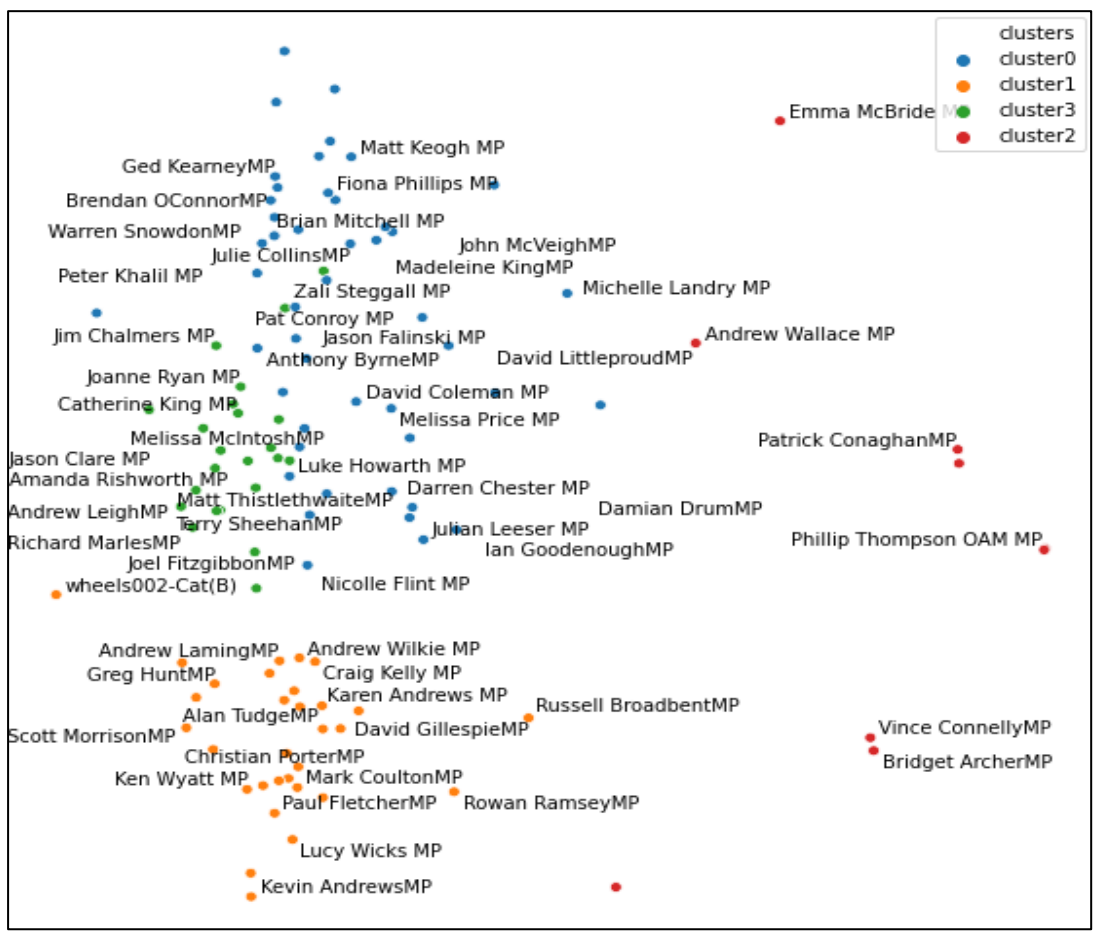

*Figure 10: Clustering analysis of the constructed KG*

To visualise the implemented KG embeddings in 3D dimension, TensorBoard[20] toolkit is used to project the resultant embeddings into low dimensional space using computed Principal Component Analysis (PCA). PCA is a statistical analyser tends to minimise dimensionality of complex problem and explore patterns from the dataset by building a linear, multivariate model from the dataset [82]. With the use of PCA, dashboard of TensorBoard provides an interesting 3D view of our KG Embeddings. Figure 11- (A, B and C) show some visuals obtained from TensorBoard. The figures illustrate the importance of visualisation to provide a subjective assessment of the implemented approach. For example, Figure 11- (A), listed all concepts related to "politics" in the high-dimensional space. Figure 11- B displays the cohort embedded with "Amanda Rishworth MP", a member of ALP. It is evident the system is able to assemble members who share the common semantic features near to each other in the embedded space. This can also be supported by members appear with "Alan Tudge MP", a member of LP, as depicted Figure 11- C. Using visualisation technique, as a subjective assessment to the implemented embedding approach, verifies the applicability and utility of our approach on constructing high quality and trustworthy domain-based KG.

---

[20] https://www.tensorflow.org/tensorboard

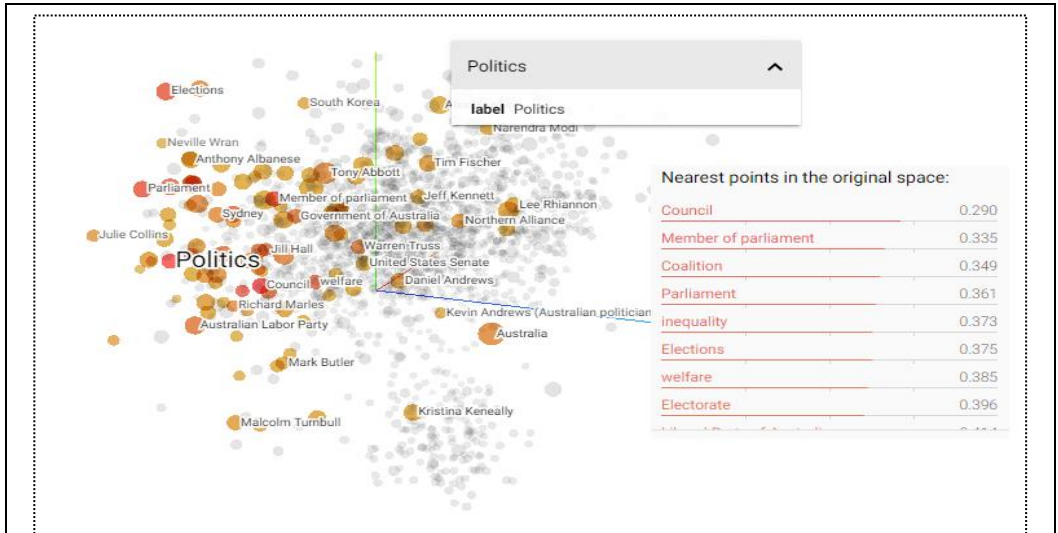

*Embedding Visualisation (A): shows concepts appear closely to "Politics" entity in same semantic vector space*

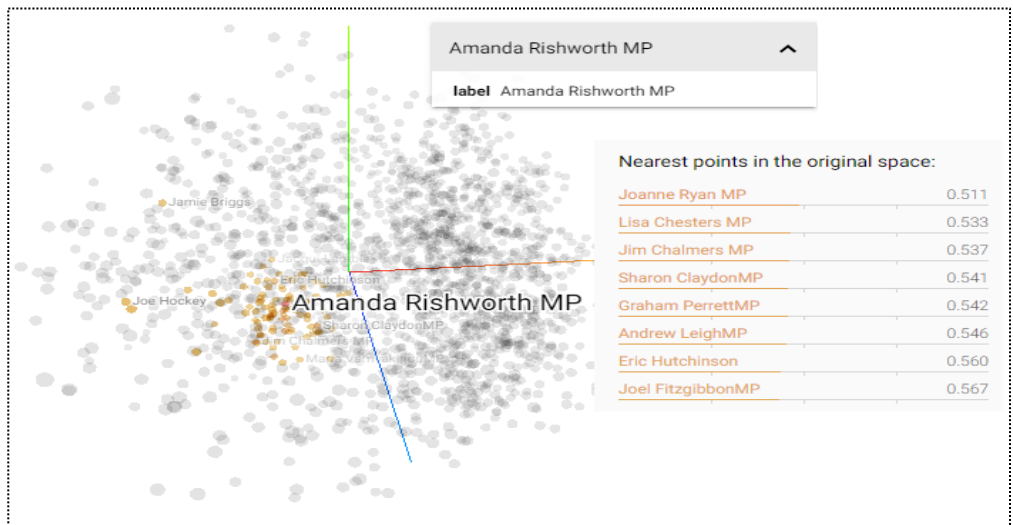

*Embedding Visualisation (B): shows members of parliament who appear in the same semantic space with Amanda Rishwoth MP*

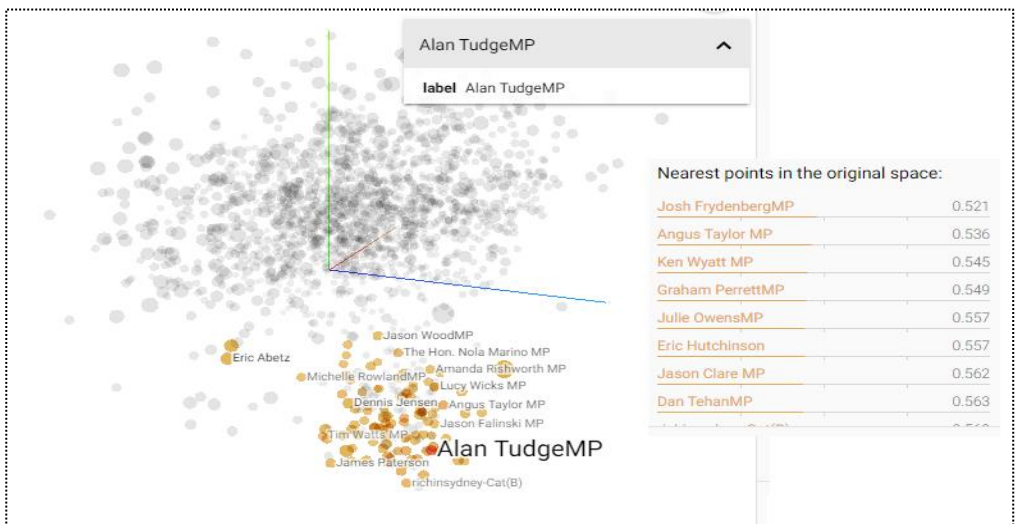

*Embedding Visualisation (C): shows members of parliament who appear in the same semantic space with Alan Tudge MP*

*Figure 11: KG Embedding visualisation using TesorBoard*

## 5. Conclusion and Future Work

The tremendous amount of information on the Web that is presented in dissimilar formats and covering various topics poses a challenge on possibilities to obtain the hoped-for added value from such massive data islands. This offers researchers a vital opportunity to consolidate efforts toward better understanding and analysis of such multimodal contents. In this context, Knowledge Graphs (KGs) are a popular phenomenon which have established a new venue by facilitating machines to understand meanings, thereby shrinking the semantic gap between them and humanity. Further, domain-based KGs have extended these exerts by propagating knowledge in dissimilar domains that can be incorporated to resolve a variety of real-life problems. Yet, the credibility of knowledge is commonly neglected in the construction of KGs especially when the knowledge is harvested from social media where spammers and other low trustworthy users find a fertile medium to publish and spread their content taking advantage of the open environment and fewer restrictions of these platforms.

This paper proposes a credibility-based domain-specific KG Embedding framework. This framework involves capturing real-life entities obtained from the social data into a formal representation depicted by domain ontology. The proposed framework embodies also a credibility module to ensure data quality and trustworthiness. The constructed KG is then embedded in low dimensional vector space using several embedding techniques. Various state-of-the-art embedding models are implemented, and their performance is evaluated using key evaluation metrics. The utility of the constructed KG Embeddings is demonstrated and substantiated on link prediction, clustering, and visualisation tasks. In the future, more embedding techniques will be implemented, and their evaluation will be studied. Also, more dimensions will be added to better framing of politics domain. Also, social data will be further scrutinised and enhanced to embody for example the sentiments of the social contents, political polarisation, etc. Another important consideration is to address the temporal factor when analysing social data; users' behaviours may change over time which affects their social credibility accordingly; hence, the temporal factor should be assimilated.